\documentclass[journal,twoside,web]{ieeecolor}
\usepackage{etoolbox}
\makeatletter
\@ifundefined{color@begingroup}%
{\let\color@begingroup\relax
\let\color@endgroup\relax}{}%
\def\fix@ieeecolor@hbox#1{%
\hbox{\color@begingroup#1\color@endgroup}}
\patchcmd\@makecaption{\hbox}{\fix@ieeecolor@hbox}{}{\FAILED}
\patchcmd\@makecaption{\hbox}{\fix@ieeecolor@hbox}{}{\FAILED}
\usepackage{tmi}
\usepackage{cite}
\usepackage{mathtools}
\usepackage{txfonts}
\usepackage{amsmath,amssymb}
\usepackage{algorithmic}
\usepackage{graphicx}
\usepackage{textcomp}
\usepackage{array}
\usepackage{booktabs} 
\usepackage{makecell}
\usepackage{multirow}
\usepackage{algorithm}
\usepackage{algorithmic}

\begin{document}
\title{SamRobNODDI: Q-Space Sampling-Augmented Continuous Representation Learning for Robust and Generalized NODDI}
\author{Taohui Xiao, Jian Cheng, Wenxin Fan, Enqing Dong, Hairong Zheng, Shanshan Wang
\thanks{This research was partly supported by the National Natural Science Foundation of China (62222118, U22A2040, 62171261, 81671848, 81371635), Shenzhen Science and Technology Program
(RCYX20210706092104034), Beijing Natural Science Foundation (L242038), Open Research Fund of the State Key Laboratory of Cognitive Neuroscience and Learning (CNLZD2101), Youth lnnovation Promotion Association CAS, National Key R\&D Program of China (2023YFA1011400), Fundamental Research Funds for the Central Universities (China), and Innovation Ability Improvement Project of Science and Technology Small and Medium-sized Enterprises of Shandong Province (2021TSGC1028).}
\thanks{T. Xiao is with the School of Mechanical, Electrical \& Information Engineering, Shandong University, Weihai 264209, China, and with the Paul C. Lauterbur Research Center for Biomedical Imaging, Shenzhen Institutes of Advanced Technology, Chinese Academy of Sciences, Shenzhen 518055, China (e-mail:thx\_loloro@163.com).}
\thanks{J. Cheng is with the School of Computer Science and Engineering, Beihang University, Beijing, 100191, China, and with the State Key Laboratory of Cognitive Neuroscience and Learning \& IDG/McGovern Institute for Brain Research, Beijing Normal University, Beijing, 100191, China (e-mail: jian\_cheng@buaa.edu.cn).}
\thanks{W. Fan, H. Zheng, and S. Wang are with the Paul C. Lauterbur Research Center for Biomedical Imaging, Shenzhen Institutes of Advanced Technology, Chinese Academy of Sciences, Shenzhen 518055, China (email: sophiasswang@hotmail.com).}
\thanks{E. Dong is with the School of Mechanical, Electrical \& Information Engineering, Shandong University, Weihai 264209, China (e-mail:enqdong@sdu.edu.cn).}
\thanks{Corresponding author:Shanshan Wang, Jian Cheng, Enqing Dong.}}
\maketitle

\begin{abstract}
Neurite Orientation Dispersion and Density Imaging (NODDI) microstructure estimation from diffusion magnetic resonance imaging (dMRI) is of great significance for the discovery and treatment of various neurological diseases. Current deep learning-based methods accelerate the speed of NODDI parameter estimation and improve the accuracy. However, most methods require the number and coordinates of gradient directions during testing and training to remain strictly consistent, significantly limiting the generalization and robustness of these models in NODDI parameter estimation. In this paper, we propose a q-space sampling augmentation-based continuous representation learning framework (SamRobNODDI) to achieve robust and generalized NODDI. Specifically, a continuous representation learning method based on q-space sampling augmentation is introduced to fully explore the information between different gradient directions in q-space. Furthermore, we design a sampling consistency loss to constrain the outputs of different sampling schemes, ensuring that the outputs remain as consistent as possible, thereby further enhancing performance and robustness to varying q-space sampling schemes. SamRobNODDI is also a flexible framework that can be applied to different backbone networks. To validate the effectiveness of the proposed method, we compared it with 7 state-of-the-art methods across 18 different q-space sampling schemes, demonstrating that the proposed SamRobNODDI has better performance, robustness, generalization, and flexibility.
\end{abstract}


\section{Introduction}

\label{sec:introduction}
\IEEEPARstart{D}{iffusion} MRI (dMRI) is a widely used medical imaging tool and the only non-invasive method for probing tissue microstructure based on the restricted diffusion of water molecules in biological tissues\cite{le1986mr}. Various dMRI models have been developed to characterize specific microstructural features, such as diffusion tensor imaging (DTI)\cite{le2001diffusion}, diffusion kurtosis imaging (DKI)\cite{steven2014diffusion}, stretched-exponential model (SEM)\cite{bennett2003characterization}, intravoxel incoherent motion (IVIM) \cite{le1988separation}, spherical mean technique (SMT)\cite{kaden2016multi}, neurite orientation dispersion and density imaging (NODDI)\cite{zhang2012noddi}, and soma and neurite density imaging (SANDI)\cite{palombo2020sandi}. Among these, NODDI is a popular microstructural model based on physiological components. NODDI-derived brain microstructural parameters can reflect changes associated with various neurological and psychiatric disorders, as well as brain development, maturation, and aging across the lifespan\cite{greenspan2016guest,fu2020microstructural}. 

Microstructural parameter estimation in NODDI is highly nonlinear and require dense q-space sampling to achieve high-quality parameter estimation. However, dense sampling necessitates extensive acquisition time and increases the susceptibility to motion artifacts\cite{zheng2022adaptive}. Additionally, the original NODDI method requires significant processing time to fit parameters using nonlinear least squares\cite{zhang2012noddi}, which limits the possibility of real-time processing on scanners\cite{park2021diffnet}. To address this limitation, various methods have been proposed to reduce computational costs \cite{daducci2015accelerated, hernandez2019using}. However, these methods remain relatively slow in processing \cite{park2021diffnet}, and their estimation performance may deteriorate when fewer directions are utilized \cite{golkov2016q, ye2017estimation, ye2019deep}. Therefore, further efforts are needed to reduce processing time while maintaining accuracy. 

Recently, deep learning (DL) methods have been widely applied to MRI data reconstruction and processing. In the field of dMRI, many studies have successfully applied deep learning to microstructural estimation, achieving the goal of reducing processing time while maintaining accuracy\cite{golkov2016q, ye2017estimation, ye2019deep, ye2020improved, zheng2023microstructure, tian2020deepdti,gibbons2019simultaneous, kebiri2024deep, chen2023deep, chen2020estimating, yang2023towards, chen2022hybrid, koppers2019spherical, sedlar2021spherical, nath2021dw, xiao2024robnoddi, caranova2023systematic}. Compared with traditional methods, deep learning-based methods can speed up estimation and maintain accuracy, but have poor generalization \cite{park2021diffnet}. Existing deep learning-based NODDI parameter estimation methods can be roughly divided into two categories, one that only considers DWI signals \cite{golkov2016q, ye2019deep, ye2020improved, zheng2023microstructure, tian2020deepdti, gibbons2019simultaneous,ye2017estimation, kebiri2024deep, chen2023deep} and the other that considers both DWI signals and b-vector information \cite{chen2020estimating, chen2022hybrid, yang2023towards, koppers2019spherical, sedlar2021spherical, nath2021dw, park2021diffnet, xiao2024robnoddi}. We describe them separately below.
\subsection{DL estimation using only DWI signals}
 We first introduce some deep learning methods for NODDI parameter estimation that use only DWI signals \cite{golkov2016q, ye2019deep, ye2020improved, zheng2023microstructure, tian2020deepdti, gibbons2019simultaneous,ye2017estimation, kebiri2024deep, chen2023deep}. For example, Golkov et al.\cite{golkov2016q} proposed using a multilayer perceptron (MLP) to estimate tissue microstructure from advanced signal models by directly mapping diffusion signals obtained from a reduced number of diffusion gradients to tissue microstructure measures. As diffusion signals can be interpreted as measurements in the q-space, this mapping was referred to as the q-space deep learning\cite{golkov2016q}. Ye et al. introduced MEDN\cite{ye2017estimation}, an end-to-end deep network that learns NODDI microstructure using a dictionary-based framework. Subsequently, Ye et al. proposed MESC-Net\cite{ye2019deep} and MESC-SD\cite{ye2020improved}, which use LSTM to estimate NODDI microstructural parameters by processing dMRI signals as 3D patches and flattening them into one-dimensional vectors for training. In a study by Gibbons et al.\cite{gibbons2019simultaneous}, a convolutional neural network with residual connections was used to simultaneously generate NODDI and GFA parameter maps from undersampled q-space imaging, achieving a 10-fold reduction in scan time compared to traditional methods. Additionally, inspired by the superior feature extraction capabilities of transformers over convolutional structures, Zheng et al.\cite{zheng2023microstructure} proposed a transformer-based learning framework called Microstructure Estimation Transformer for Sparse Coding (METSC) for microstructure parameter estimation. In summary, these deep learning methods for microstructural parameter estimation have achieved good performance, enabling accelerated parameter estimation while ensuring accuracy. However, these methods only consider DWI signals, requiring the number and coordinates of the directions during testing to be strictly consistent with those during training. This leads to poor generalization and clinical applicability, limiting their use in real clinical settings.
\subsection{DL estimation using both DWI signals and b-vectors}
Additionally, there are methods that take both the DWI signal and b-vector information into account simultaneously \cite{chen2020estimating, chen2022hybrid, yang2023towards, koppers2019spherical, sedlar2021spherical, nath2021dw, park2021diffnet, xiao2024robnoddi}. For example,  Chen et al.\cite{chen2020estimating} proposed a graph convolutional neural network (GCNN)-based method for estimating NODDI parameter from undersampled dMRI, which fully considers q-space angular neighboring information. Furthermore, Chen et al.\cite{chen2022hybrid} introduced a hybrid graph transformer (HGT) method, which explicitly considers q-space geometry using graph neural networks (GNN) and leverages spatial information through a novel residual dense transformer (RDT). To fully utilize both 3D spatial and angular information in dMRI signals, Chen et al.\cite{yang2023towards} further proposed 3D HGT to enhance NODDI estimation performance. Park et al.\cite{park2021diffnet} proposed a generalized diffusion parameter mapping network (DIFFnet), which projects diffusion data into a fixed-size matrix called "Qmatrix" to be used as input for training. The Qmatrix incorporates the gradient directions and b-value information for each signal, allowing for variable gradient schemes and b-values during actual testing. However, discretized projection may lead to a loss of precision. Additionally, higher dimensions result in slower processing speeds. Moreover, this method is voxel-wise, operating on a per-voxel basis without considering the redundant information between patches. Vishwesh et al.\cite{nath2021dw} directly used SH coefficients to estimate microstructural parameters. By converting the DWI signal into an SH representation with a fixed dimensionality, different microstructural parameters can be estimated with variable numbers of diffusion directions. However, this study focuses on using single-shell DWI signals with all directions to estimate different microstructural parameters, without adequately considering multi-shell DWI signals. Additionally, the information between different sampling directions is not fully exploited. Sedlar et al.\cite{sedlar2021spherical} applied spherical convolutional neural networks (CNNs) to white matter NODDI microstructural imaging in dMRI, proposing a spherical CNN model with full-spectrum convolution and nonlinear layers (Fourier\_s2cnn). FourierS2cnn uses dMRI signals as input, followed by a denoising layer, and then converts them into SH coefficients for training. Since the denoising layer directly operates on the dMRI signals, the performance significantly degrades when the sampling directions during testing differ from those used in training, thus failing to handle the generalization of sampling directions.

To address these issues, we propose a q-space sampling augmentation-based continuous representation learning framework for robust and generalized NODDI. Our main contributions can be summarized as follows:
\begin{itemize}
\item A q-space sampling-augmented continuous representation learning framework (SamRobNODDI) is developed to achieve robust and generalized NODDI.
\item A novel continuous representation learning method with q-space sampling augmentation is proposed to fully explore the information across different diffusion directions, thereby improving robustness and flexibility to varying diffusion directions.
\item A sampling consistency loss is designed to constrain output results across different sampling schemes, further enhancing the performance and robustness of the proposed method.
\item Extensive experimental validation is conducted, including experimental comparisons (different sampling schemes in training and testing, different sampling rates, etc.) and ablation studies (different losses, different backbones, etc.). The results demonstrate that our proposed SamRobNODDI exhibits better performance, robustness, and flexibility compared to existing state-of-the-art methods.
\end{itemize}

The rest of the paper is organized as follows. Section \ref{sec2} describes the proposed framework in details. Section \ref{sec3} presents experimental setup. Section \ref{sec4} demonstrates the results and discussion. Finally, Section \ref{sec5} provides the conclusions of the paper.

\begin{figure*}[!t]
\centerline{\includegraphics[width=\textwidth]{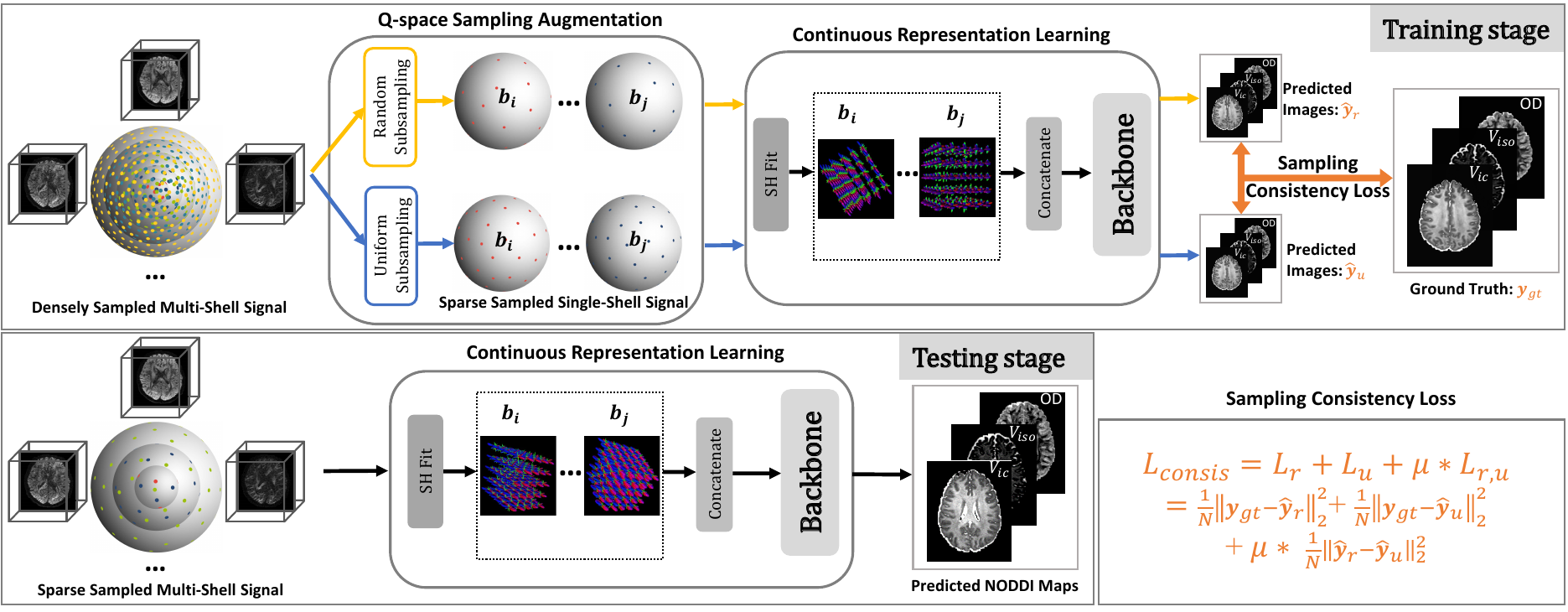}}
\caption{Overview of SamRobNODDI. It contains both a training stage and a testing stage. During the training stage, the process consists of q-space sampling augmentation, continuous representation learning, and consistency loss. In the testing stage, only the continuous representation learning process is performed.}
\label{fig1}
\end{figure*} 

\section{METHODS}
\label{sec2}
In this section, we provide a detailed description of the proposed method, including the overall framework of SamRobNODDI, q-space sampling augmentation, continuous representation learning, and consistency loss.

\subsection{SamRobNODDI}
Figure 1 shows the overall framework of our proposed method. After training, our method does not require the sampling scheme in testing to be consistent with that in training. Our method is applicable to both 2D slices and 3D patches. In the training phase, we introduce a q-space sampling augmentation module, which includes two types of subsampling: uniform subsampling and random subsampling. We assume that the input full-direction data has dimensions of ($x, y, z, q$). The dimension for random subsampled data is ($x, y, z, q_{r}$), while the dimension for uniform subsampled data is ($x, y, z, q_{u}$). A detailed description of random sampling and uniform sampling will be presented in Section B. After performing sampling augmentation on the DWI signal, we obtain two subsampled DWI images. These images are then for continuous representation learning. We perform sixth-order SH fitting on the two downsampled DWI images separately, obtaining coefficients with dimensions of ($x, y, z, c$). After concatenation, the coefficient dimensions become ($x, y, z, nc$), where n represents the number of shells used. The concatenated SH coefficients are then fed into the backbone network. A detailed description of continuous representation learning will be presented in Section C. Since we have two types of inputs, we obtain two outputs, which are then connected to the ground truth using a sampling consistency loss. SamRobNODDI takes b-vector information into account and aims to enable our model to fully exploit and utilize the information from DWI signals in different sampling directions. In the testing phase, we directly use the trained model to estimate the sampled data, without requiring the sampling scheme to match that used during training, resulting in robust NODDI parameter outputs.

It should be noted that our proposed method is independent of the backbone network, offering great flexibility to adopt any neural network as the backbone. Due to its unique residual connections, ResNet has shown excellent performance in medical image analysis tasks. Therefore, in this study, we use a modified ResNet18\cite{he2016deep} as the base architecture. For ResNet18, we remove the original pooling layers, batch normalization layers, and the final fully connected layer, retaining only the convolutional layers, activation layers, and residual connections.

\subsection{Q-space Sampling Augmentation}
In dMRI research, uniform sampling is commonly used to obtain more information with fewer sampling directions\cite{caruyer2013design,cheng2014designing,cheng2017single}. In our study, we primarily focus on enhancing the model's generalization across different sampling schemes, aiming for consistent performance regardless of the sampling method. To achieve this, we propose a q-space sampling augmentation module that simultaneously applies two types of subsampling: random subsampling and uniform subsampling.

Before training, we perform uniform subsampling on the full sampling data with $q$ directions ($x, y, z, q$), obtaining a subsampled image with $q_{u}$ dimensions ($x, y, z, q_{u}$). The uniform sampling is implemented using the method described in \cite{cheng2017single}. For random sampling, we randomly select qr directions from the $q$ directions, resulting in an image with dimensions ($x, y, z, q_{r}$), where $q_{r}$ varies for each sample in every iteration, meaning both the number of directions and their coordinates change.By using sampling augmentation module, we enable the model to adapt to different sampling schemes, thereby improving its robustness to variations in sampling directions.

\subsection{Continuous Representation Learning}
Spherical Harmonics (SH) are commonly used in dMRI\cite{koppers2019spherical,sedlar2021spherical,nath2021dw}, due to their properties of orthogonality and continuity. Typically, the dMRI signal can be represented as a linear combination of spherical harmonic basis elements, as shown in Equation (1):
\begin{equation}
\mathrm{\mathbf{E}}(u)=\sum_{l=0}^{\infty} \sum_{m=-l}^{l} \mathbf{c}_{l}^{m} \mathbf{B}_{l}^{m}(u),
\end{equation}where, $\mathrm{\mathbf{E}}(u)$ represents the normalized dMRI signal, $u$ represents a unit-norm vector, $\mathbf{c}_{l}^{m}$  represents the SH coefficient of dMRI, $\mathbf{B}_{l}^{m}(u)$ represents the SH basis. The SH basis is formed by gradient direction table through the utilization of Legendre polynomials. $l$ and $m$ represent SH order and degree respectively. We use a least squares method with a smoothing constraint term \cite{descoteaux2007regularized} to calculate the SH coefficient, with the objective function given by:

\begin{equation}
\min _{\mathbf{c}}\|\mathbf{E}-\mathbf{B} \mathbf{c}\|_{2}^{2}+\lambda\|\mathbf{L} \mathbf{c}\|_{2}^{2},
\end{equation}where $\mathbf{E}$ is the normalized vector representation of the DWI signals, $\mathbf{B}$ is the corresponding SH basis function matrix, $\mathbf{c}$ is the vector of SH coefficients we want to estimate, $\lambda$ is the weight hyperparameter for the smoothing term, and $\mathbf{L}$ is the Laplace–Beltrami matrix \cite{descoteaux2007regularized}. By solving the above least squares problem, we can obtain the estimated SH coefficients as follows:

\begin{equation}
\mathbf{c}=\left(\mathbf{B}^{T} \mathbf{B}+\lambda \mathbf{L}^{T} \right)^{-1}\left(\mathbf{B}^{T} \mathbf{E}.\right)
\end{equation}

We set $\lambda$ to 0.006 as described in \cite{descoteaux2007regularized}. Different subsampled DWI signals yield different SH coefficients. That is, for the random subsampling data ($x, y, z, q_{r}$) and the uniform subsampling data ($x, y, z, q_{u}$), the fitted SH coefficients will differ. Therefore, fitting only SH coefficients from uniform subsampling cannot fully exploit the directional information of the DWI signal, making it necessary to perform sampling augmentation before conducting SH fitting.

We combine sampling augmentation with continuous representation learning, allowing us to incorporate different b-vector information. This approach facilitates a thorough exploration of the relationships between signals obtained from different sampling schemes and ensures that the model learns effectively from the fitted SH coefficients. As a result, regardless of the sampling direction, the estimated outcomes will remain as consistent as possible, enhancing the model's robustness and generalization. Additionally, the smoothness constraint is important for fitting the SH coefficients, as it can reduce the impact of noise and provide a robust estimate with a limited number of samples.

\subsection{Sampling Consistency Loss}
For the $n$-th training sample $\mathbf{x}_{n}$, we have a paired input-output patch or slice ($\mathbf{x}_{n}$, $\mathbf{y}_{n}$). For uniform subsampling, we obtain an estimated output $\hat{\mathbf{y}}_{n}^{u}$, and for random subsampling, we obtain an estimated output $\hat{\mathbf{y}}_{n}^{r}$. Based on the target $\mathbf{y}_{n}$ and the two different outputs $\hat{\mathbf{y}}_{n}^{u}$ and $\hat{\mathbf{y}}_{n}^{r}$, we can define three mean squared error MSE losses as follows:
\begin{equation}
L_{r}=\frac{1}{N} \sum_{n=1}^{N}\left\|\mathbf{y}_{n}- \hat{\mathbf{y}}_{n}^{r}\right\|_{\mathrm{F}}^{2}.
\label{loss}
\end{equation}

\begin{equation}
L_{u}=\frac{1}{N} \sum_{n=1}^{N}\left\|\mathbf{y}_{n}-\hat{\mathbf{y}}_{n}^{u}\right\|_{\mathrm{F}}^{2}.
\label{loss}
\end{equation}

\begin{equation}
L_{u}=\frac{1}{N} \sum_{n=1}^{N}\left\|\hat{\mathbf{y}}_{n}^{r}-\hat{\mathbf{y}}_{n}^{u}\right\|_{\mathrm{F}}^{2}.
\label{loss}
\end{equation}

Therefore, we define the sampling consistency loss as follows:
\begin{equation}
\begin{aligned}
L_{consis}=& L_{r}+L_{u}+\mu*L_{r,u}=\frac{1}{N} \sum_{n=1}^{N}\left\|\mathbf{y}_{n}- \hat{\mathbf{y}}_{n}^{r}\right\|_{\mathrm{F}}^{2}+\\& \frac{1}{N} \sum_{n=1}^{N}\left\|\mathbf{y}_{n}-\hat{\mathbf{y}}_{n}^{u}\right\|_{\mathrm{F}}^{2} + \mu*\frac{1}{N} \sum_{n=1}^{N}\left\|\hat{\mathbf{y}}_{n}^{r}-\hat{\mathbf{y}}_{n}^{u}\right\|_{\mathrm{F}}^{2}
\end{aligned},\\
\end{equation}
where $\mu$ is a hyperparameter used to adjust the weight of the loss between the outputs of random subsampling and uniform subsampling. 

By using the sampling consistency loss, we impose a constraint on the outputs of different subsampled DWI signals, which helps further improve the performance and robustness of the proposed method across different sampling schemes.

\section{Experiments}
\label{sec3}
In this section, we introduce the dataset, comparison methods, and Implementation Details used in the paper.

\subsection{Dataset}
We randomly selected 50 subjects from the Human Connectome Project (HCP) dataset\cite{van2013wu} for training and testing. Among them, 30 subjects were used for training and 20 subjects for testing. Additionally, 3 subjects were randomly selected from the HCP dataset for validation. The dMRI scans of these subjects were obtained using three b-values: b = 1000, 2000, and 3000 $s/{mm}^2$. Each b-value includes 90 gradient directions, and the image resolution is 1.25mm isotropic. The training and ground truth microstructure images were obtained using the AMICO algorithm \cite{daducci2015accelerated}, utilizing all 270 gradient directions. Compared to the original NODDI model fitting, AMICO significantly reduces the computation time without sacrificing estimation quality\cite{daducci2015accelerated}, and it has been incorporated into the standard processing pipeline for the UK Biobank dataset\cite{miller2016multimodal}. 
\begin{table*}[ht]
\centering
\caption{Quantitative indicators of NODDI parameters for SS testing using different methods on the q-space data with 30 diffusion directions per shell (1000, 2000 $s/{mm}^2$).} 
\label{tab1}
\resizebox{\textwidth}{!}{
\tiny 
\begin{tabular}{c|cccc|clll}
\hline
\multirow{2}{*}{Method}                                                                 & \multicolumn{4}{c|}{PSNR}                  & \multicolumn{4}{c}{SSIM}               \\   \cline{2-9}                & $V_{ic}$      & $V_{iso}$      & OD       & All      & $V_{ic}$       & \multicolumn{1}{c}{$V_{iso}$}    & \multicolumn{1}{c}{OD}      & \multicolumn{1}{c}{All}     \\ \hline
AMICO\cite{daducci2015accelerated}                                                                 
                & 21.55  $\pm$ 1.11              & 32.23  $\pm$ 0.59       & 23.70  $\pm$ 0.55             & 24.02   $\pm$ 0.82              & 0.9265  $\pm$ 0.0135  & 0.9772    $\pm$ 0.0039              & 0.9460     $\pm$ 0.0065           & 0.9499  $\pm$ 0.0067  \\ \hline
q-DL\cite{golkov2016q}                                                                   
                & 28.33  $\pm$ 0.68              & 33.29  $\pm$ 0.68       & 26.52  $\pm$ 0.48             & 28.57   $\pm$ 0.54              & 0.9747  $\pm$ 0.0034  & 0.9809    $\pm$  0.0038              & 0.9603     $\pm$ 0.0047           & 0.9720  $\pm$ 0.0036  \\ \hline

U-Net++\cite{zhou2018unet++}                                                                            & 31.67  $\pm$ 0.69               & 36.32  $\pm$ 0.65               & 29.33  $\pm$ 0.46              & 31.58   $\pm$ 0.49              & 0.9760  $\pm$ 0.0016  & 0.9822  $\pm$ 0.0022                 & 0.9764    $\pm$ 0.0028             & 0.9782 $\pm$ 0.0020   \\ \hline

CNN\cite{gibbons2019simultaneous}                                                                      & 32.39  $\pm$ 0.62               & 36.98 $\pm$ 0.64                & 30.10  $\pm$ 0.48              & 32.32   $\pm$ 0.49              & 0.9879  $\pm$ 0.0017  & 0.9905  $\pm$ 0.0019                 & 0.9818   $\pm$ 0.0023              & 0.9867  $\pm$ 0.0018  \\ \hline

MESC-SD\cite{ye2020improved}                                                                & 32.10   $\pm$ 0.64              & 37.13     $\pm$ 0.64             & 30.27  $\pm$ 0.50             & 32.33    $\pm$ 0.51              & 0.9878    $\pm$ 0.0017   & 0.9911   $\pm$ 0.0017                & 0.9831   $\pm$ 0.0023             & 0.9874    $\pm$ 0.0017                         \\ \hline

Fourier\_s2cnn\cite{sedlar2021spherical}                                                                     & 29.88  $\pm$ 0.65             & 35.46   $\pm$ 0.77              & 28.84    $\pm$ 0.60            & 30.59    $\pm$ 0.62              & 0.9818   $\pm$ 0.0030  & 0.9867   $\pm$ 0.0028                & 0.9773    $\pm$ 0.0036             & 0.9819   $\pm$ 0.0029  \\ \hline

HGT\cite{chen2022hybrid}                                                                              & 32.60   $\pm$ 0.63           & 37.30 $\pm$ 0.67                & 30.33   $\pm$ 0.48             & 32.56     $\pm$ 0.50             &0.9886  $\pm$ 0.0016   & 0.9911  $\pm$ 0.0016               & 0.9829   $\pm$ 0.0022             & 0.9875  $\pm$ 0.0017        \\ \hline

SamRobNODDI      & \textbf{32.61}  \textbf{$\pm$ 0.66}              & \textbf{37.36}   \textbf{$\pm$ 0.64}               & \textbf{30.41}   \textbf{$\pm$ 0.45}             & \textbf{32.61}   \textbf{$\pm$ 0.48}               & \textbf{0.9887}  \textbf{$\pm$ 0.0016}     & \textbf{0.9913}   \textbf{$\pm$ 0.0016}                & \textbf{0.9834}   \textbf{$\pm$ 0.0021}             & \textbf{0.9878}     \textbf{$\pm$ 0.0016}                     \\  \hline
\end{tabular}}
\end{table*}
\begin{figure*}[!t]
\centerline{\includegraphics[width=0.85\textwidth]{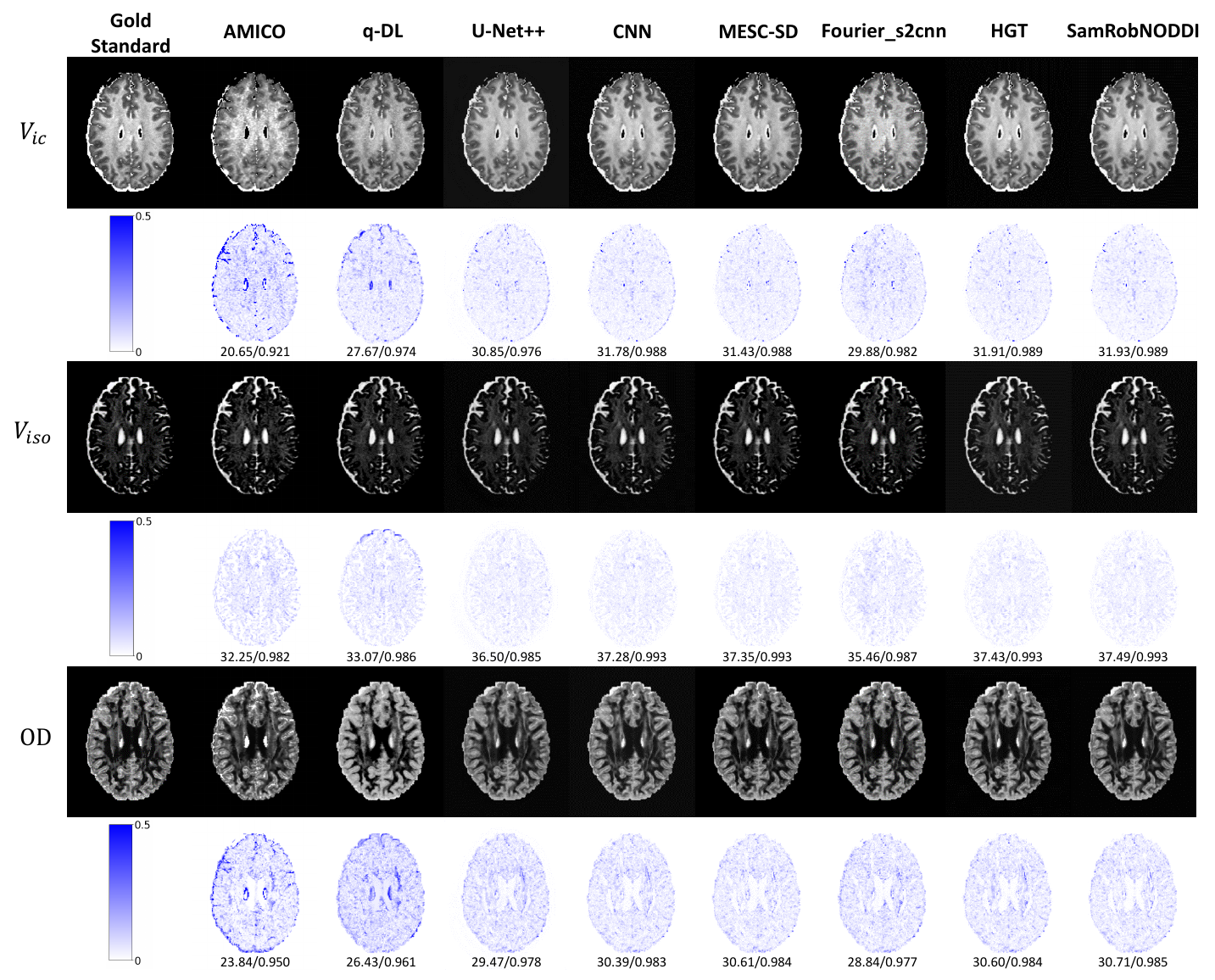}}
\caption{Visualization results of NODDI parameter for SS testing using different
methods with 30 diffusion directions per shell (1000, 2000 $s/{mm}^2$).}
\label{fig2}
\end{figure*} 

\subsection{Compared Methods}
Our comparative methods include one traditional method: AMICO\cite{daducci2015accelerated}; four deep learning methods that consider only DWI signals: q-DL\cite{golkov2016q}, U-Net++\cite{zhou2018unet++}, 2D-CNN\cite{gibbons2019simultaneous}, MESC-SD\cite{ye2020improved}, and HGT\cite{chen2022hybrid}; and two methods that simultaneously considers both DWI signals and b-vector information: Fourier\_s2cnn\cite{sedlar2021spherical}. AMICO\cite{daducci2015accelerated} is a traditional method used to accelerate NODDI estimation. q-DL\cite{golkov2016q} is a q-space deep learning method based on MLP. U-Net++ is implemented as described in the literature\cite{zhou2018unet++,chen2022hybrid}. 2D-CNN\cite{gibbons2019simultaneous} is a 2D CNN network with two residual connections proposed by Gibbons et al. for microstructure estimation. MESC-SD\cite{ye2020improved} uses LSTM for NODDI microstructure estimation. Fourier\_s2cnn\cite{sedlar2021spherical} is a method based on spherical CNNs with rotational invariance. HGT\cite{chen2022hybrid} is a method based on graph neural networks and transformers. Among these methods, q-DL and Fourier\_s2cnn are voxel-wise approaches, where the model input is the DWI signal from a single voxel. U-Net++, 2D-CNN, and HGT are 2D slice-wise approaches that use 2D slices of DWI signals as input. MESC-SD estimates the microstructure of the central voxel by dividing the DWI signal into 3D patches of different sizes. Considering memory size and time costs, we conducted comparative experiments on MESC-SD using patches of size $3 \times 3 \times 3$ for both input and output.
\begin{table*}[ht]
\centering
\caption{ Quantitative indicators of NODDI parameters for RS testing using different methods on the q-space data with 30 diffusion directions per shell (1000, 2000 $s/{mm}^2$).} 
\label{tab2}
\resizebox{\textwidth}{!}{
\tiny 
\begin{tabular}{c|cccc|clll}
\hline
\multirow{2}{*}{Method}                                                                 & \multicolumn{4}{c|}{PSNR}                  & \multicolumn{4}{c}{SSIM}               \\   \cline{2-9}
   & $V_{ic}$      & $V_{iso}$     & OD       & All      & $V_{ic}$      & \multicolumn{1}{c}{$V_{iso}$}    & \multicolumn{1}{c}{OD}      & \multicolumn{1}{c}{All}     \\ \hline
AMICO\cite{daducci2015accelerated}                                                                  
                & 21.52  $\pm$ 1.13              & 31.98 $\pm$ 0.81       & 22.72  $\pm$ 1.01             & 23.60   $\pm$ 0.97              & 0.9245  $\pm$ 0.0145  & 0.9753    $\pm$ 0.0050              & 0.9360     $\pm$ 0.0110           & 0.9452  $\pm$ 0.0086  \\ \hline
                
q-DL\cite{golkov2016q}                                                               & 26.85    $\pm$ 0.82             & 31.37   $\pm$ 0.77              & 19.71     $\pm$ 0.65           & 23.47   $\pm$ 0.62               & 0.9643 $\pm$ 0.0062   & 0.9695  $\pm$ 0.0072                 & 0.8651    $\pm$ 0.0171             & 0.9329   $\pm$ 0.0091    \\ \hline

U-Net++\cite{zhou2018unet++}                                                              
        & 28.09   $\pm$ 1.05              & 32.25    $\pm$ 1.00              & 17.15      $\pm$ 0.84          & 21.46   $\pm$ 0.83        & 0.9585   $\pm$ 0.0067  & 0.9639   $\pm$ 0.0080         & 0.8646   $\pm$ 0.0201        & 0.9290  $\pm$ 0.0106  \\ \hline

CNN\cite{gibbons2019simultaneous}                                                              
          & 27.54   $\pm$ 0.85             & 32.02  $\pm$ 1.00                & 16.37     $\pm$ 0.84           & 20.71    $\pm$ 0.83       & 0.9719   $\pm$ 0.0058  & 0.9727  $\pm$ 0.0073         & 0.8542    $\pm$ 0.0215       & 0.9330  $\pm$ 0.0108  \\ \hline

MESC-SD\cite{ye2020improved}                                                             
 & 28.10     $\pm$ 0.81            & 32.62    $\pm$ 0.97              & 16.94    $\pm$ 0.79           & 21.27   $\pm$ 0.75              & 0.9723  $\pm$ 0.0055   & 0.9749   $\pm$ 0.0075                & 0.8603    $\pm$ 0.0215             & 0.9358  $\pm$ 0.0097    \\ \hline

Fourier\_s2cnn\cite{sedlar2021spherical}                                                       
   & 25.88   $\pm$ 0.60             & 32.01  $\pm$ 0.71               & 16.27   $\pm$ 0.54             & 20.48   $\pm$ 0.54               & 0.9667   $\pm$ 0.0052  & 0.9728    $\pm$ 0.0061              & 0.8163   $\pm$ 0.0213                & 0.9186  $\pm$ 0.0102   \\ \hline

HGT\cite{chen2022hybrid}                                                
 & 28.65    $\pm$ 1.05             & 32.71   $\pm$ 1.06               & 17.82    $\pm$ 0.82            & 22.11    $\pm$ 0.82             & 0.9711   $\pm$ 0.0067  & 0.9734    $\pm$ 0.0074               & 0.8764  $\pm$ 0.0189             & 0.9403    $\pm$ 0.0102                       \\ \hline

SamRobNODDI
 & \textbf{32.22}     \textbf{$\pm$ 0.65}            & \textbf{36.81}    \textbf{$\pm$ 0.66}              & \textbf{30.15}      \textbf{$\pm$ 0.45}          & \textbf{32.27}    \textbf{$\pm$ 0.49}              & \textbf{0.9875}   \textbf{$\pm$ 0.0017}     & \textbf{0.9902}   \textbf{$\pm$ 0.0018}                & \textbf{0.9822}    \textbf{$\pm$ 0.0022}             & \textbf{0.9866}   \textbf{$\pm$ 0.0017}   \\ \hline
\end{tabular}}
\end{table*}
\begin{figure*}[!t]
\centerline{\includegraphics[width=0.85\textwidth]{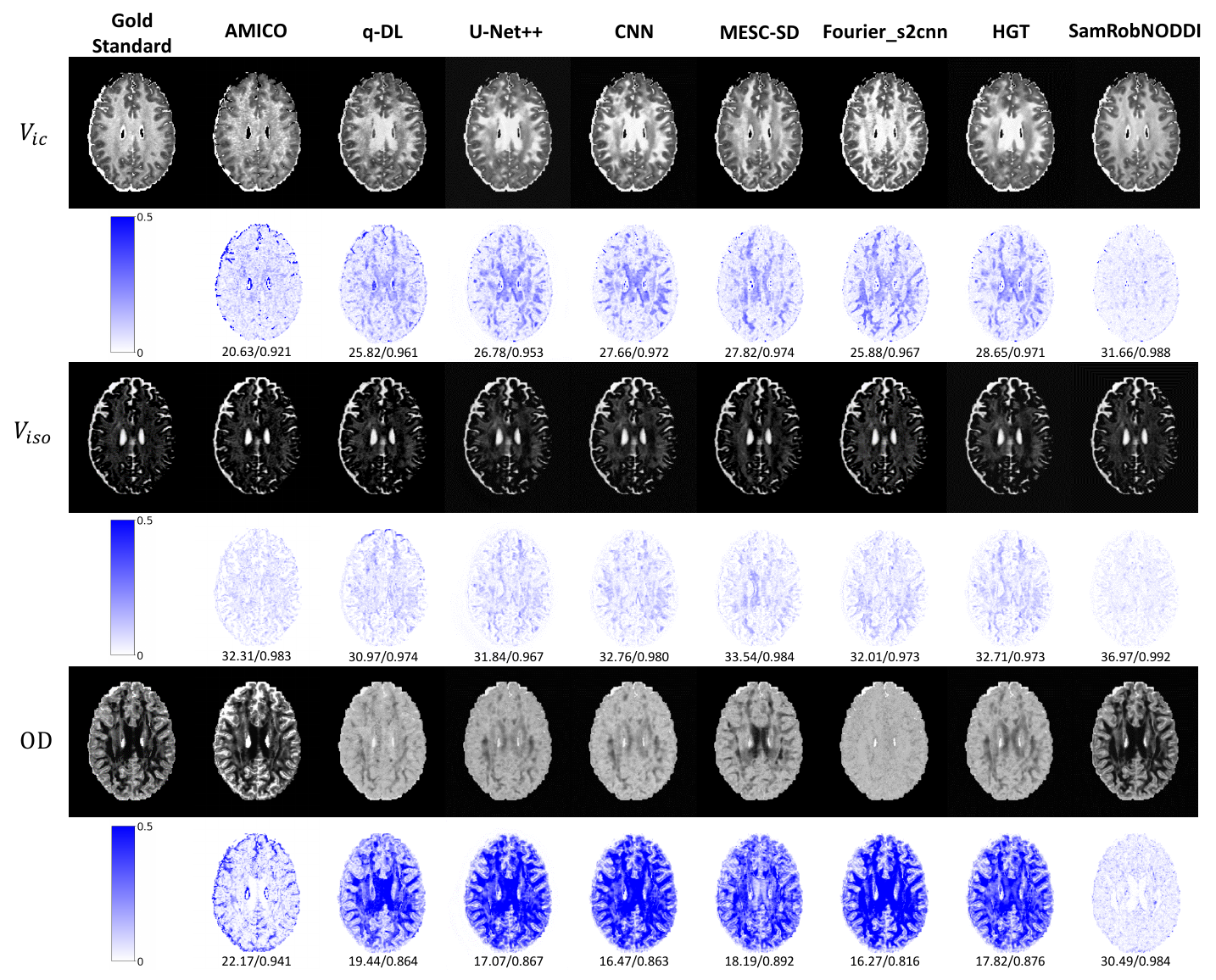}}
\caption {Visualization results of NODDI parameter for RS testing using different
methods with 30 diffusion directions per shell (1000, 2000 $s/{mm}^2$).}
\label{fig3}
\end{figure*} 
\subsection{Implementation Details}
Our experiments were processed using four NVIDIA TITAN XP GPUs with 12GB of memory each. We used the Adam optimizer, setting the initial learning rate of our method to 0.001, and for comparison methods, we used the initial learning rates specified in their original papers or open-source codes. All comparative methods use MSE as the loss function, consistent with the original paper, while our method employs sampling consistency loss as shown in Equation (7) with the parameter 
$\mu$ set to 0.001. We used sixth-order fitting for each shell in the SH coefficient fitting. We train our method using 2D slices. When comparing with MESC-SD as the backbone, we use a $3 \times 3 \times 3$ patch for training. After training, the best model was used for testing. We used the peak signal-to-noise ratio (PSNR), and structural similarity index measure (SSIM) as the evaluation metrics for all experiments.
\begin{figure*}[!t]
\centerline{\includegraphics[width=0.70\textwidth]{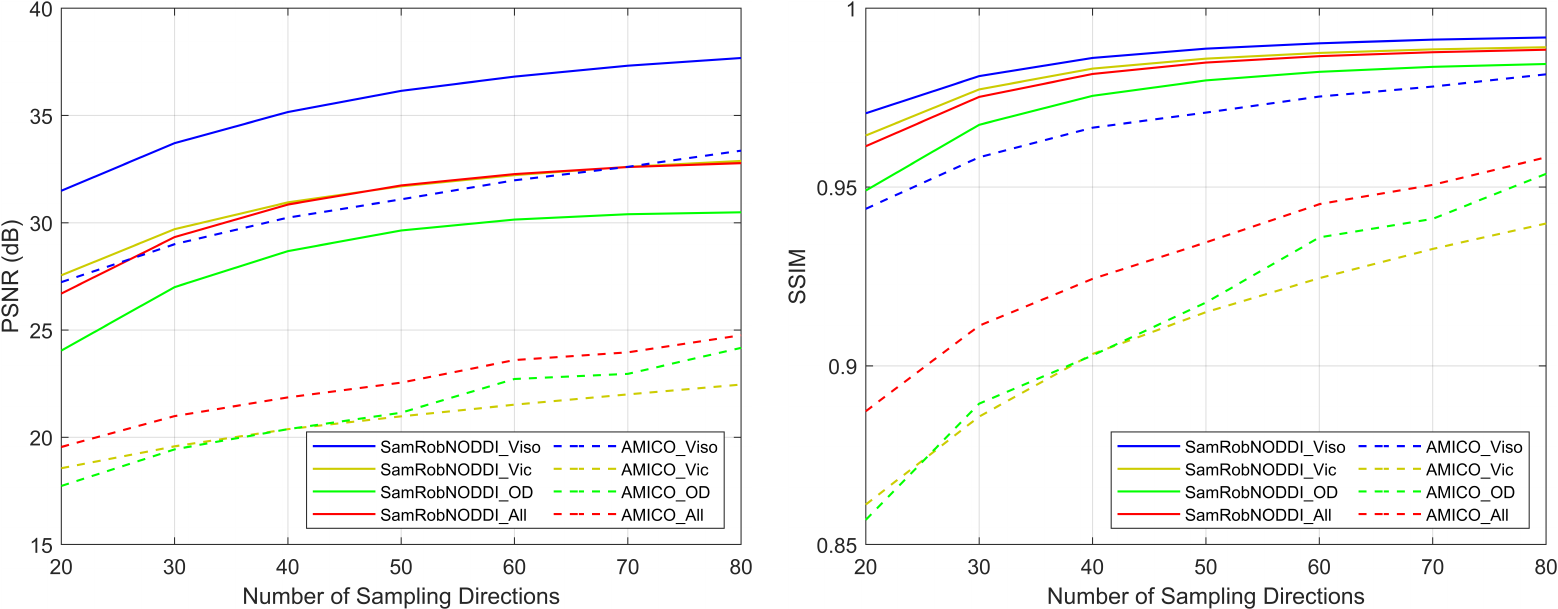}}
\caption{Comparison of PSNR and SSIM for NODDI parameters estimated by SamRobNODDI and AMICO using different sampling rates.}
\label{fig4}
\end{figure*} 

\section{RESULTS AND DISCUSSION}
\label{sec4}
In this section, we present the experimental results and discussion. Throughout all the experiments, we use \textbf{SS} (same sampling) to indicate that the same sampling scheme is used for both training and testing. \textbf{RS} (random sampling) means using a random sampling scheme during testing, which is different from uniform subsampling used in the training phase. All methods were trained using a total of 60 directions (30 directions for b=1000 $s/{mm}^2$ and 30 directions for b=2000 $s/{mm}^2$). We compare the performance and generalization ability of our method with existing state-of-the-art methods under both SS and RS testing. We also compared the results of all methods at different sampling rates. Additionally, we conducted several sets of ablation experiments to further validate the effectiveness of the proposed method.

\subsection{Comparison under SS (same sampling) testing}
We first perform SS testing, where all methods use the same sampling directions for testing as those used during training.

Table \ref{tab1} shows the average quantitative metrics and standard deviations for testing subjects, while Fig. \ref{fig2} presents a comparison of NODDI parameter visualization results for a randomly selected subject. Since only 60 directions are used for testing, it can be observed that the AMICO with fewer directions performs much worse than the full-direction AMICO estimation. Most deep learning-based methods yield reasonable estimation results. Our method does not show a noticeable difference compared to most deep learning comparative methods under SS testing, but it performs slightly better in terms of metrics.

Moreover, we can see that all methods achieve the best estimation results for the isotropic volume fraction ($V_{iso}$), followed by the intracellular volume fraction ($V_{ic}$) and the orientation dispersion index (OD). This might be because $V_{iso}$ is easier to estimate compared to the other two parameters, hence all methods achieve better results for this parameter.
\begin{figure*}[!t]
\centerline{\includegraphics[width=0.85\textwidth]{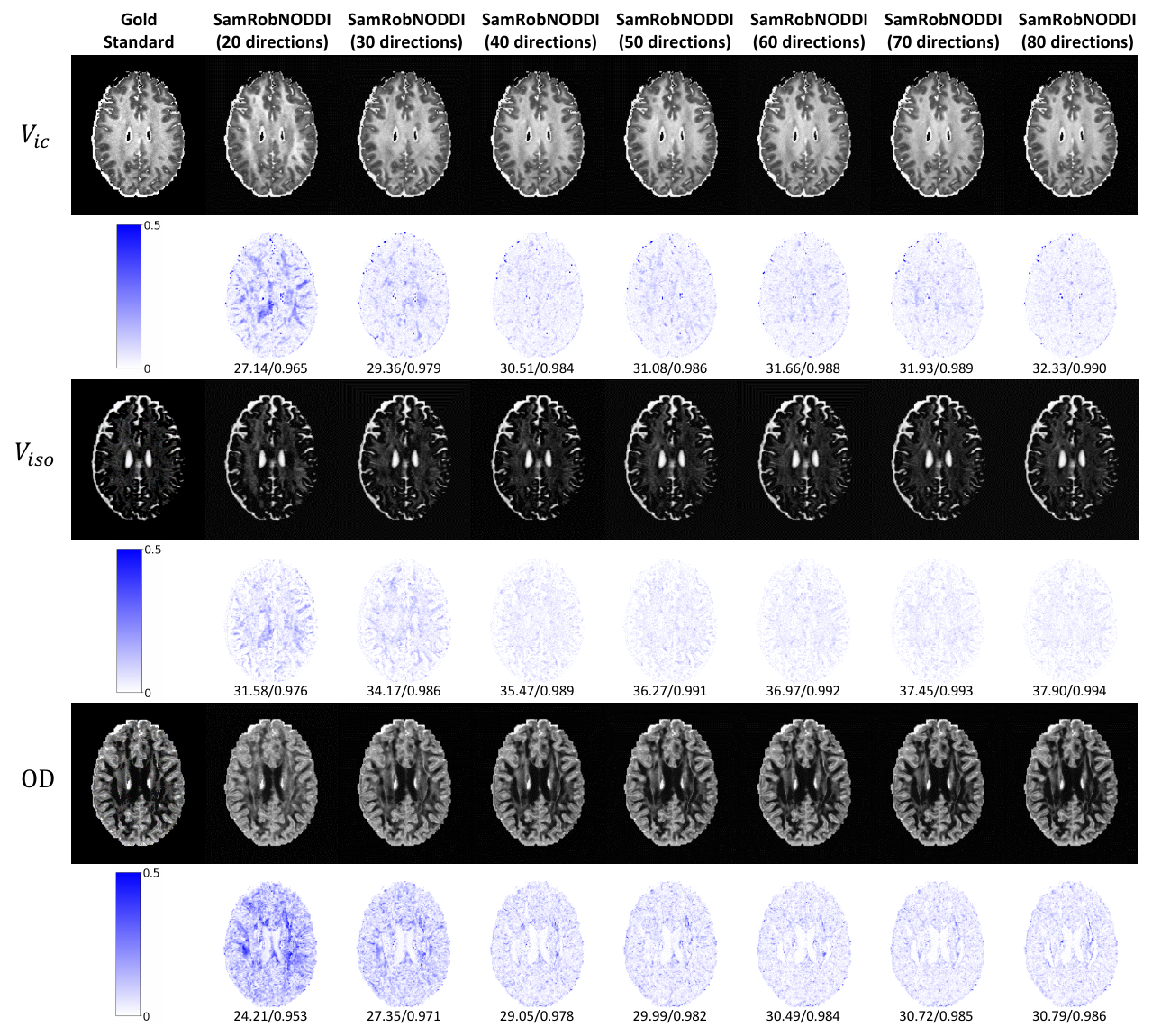}}
\caption{Visualization results of NODDI parameters estimated by SamRobNODDI using different sampling rates.}
\label{fig5}
\end{figure*} 
\subsection{Comparison under RS (random sampling) testing}
To validate the generalization of the proposed method, we conduct RS testing with existing deep learning methods based on the previously trained model, where the diffusion directions during testing differ from those used in training for all methods. Table \ref{tab2} shows the average quantitative metrics and standard deviations for 20 subjects, while Fig. \ref{fig3} presents the NODDI parameter visualization results for a randomly selected subject. From these results, it is evident that the performance of current deep learning methods significantly degrades under RS testing, making it challenging to accurately recover usable NODDI parameters. Specifically, $V_{iso}$ and $V_{ic}$ are less sensitive to diffusion gradient directions. Hence, their performance degradation is relatively lower. However, compared to SS testing, there is still a noticeable decline.

For the OD, it is more sensitive to gradient directions compared to other parameters. Therefore, when the sampling scheme changes, the impact on the OD is the greatest. As shown in Fig. \ref{fig3}, it is nearly impossible to estimate the correct results.

In contrast, the proposed method demonstrates much more stable results under RS testing, significantly outperforming existing methods. It provides reliable estimations for all three NODDI parameters, indicating robust performance against varying diffusion directions in DWI signals.


\subsection{Generalization test with different sampling rates}
To further validate the robustness and performance of the proposed method in practical applications, we used the previously trained model to estimate NODDI parameters from DWI signals with different sampling directions in q-space and compared the results with the traditional AMICO method. 
Fig. \ref{fig4} compares the PSNR and SSIM metrics estimated by our method and AMICO at different sampling rates, which we set as: 20, 30, 40, 50, 60, 70, and 80. The number of sampling directions for the two shells is divided in a 1:1 ratio. The results of the quantitative metrics show that as the sampling rates increases, the results of both methods improve, but our proposed method clearly outperforms AMICO. Furthermore, among the three parameters estimated by our trained model and AMICO, the $V_{iso}$ result is the best. Fig. \ref{fig5} presents a comparison of the visualization results of our method. It is evident that the differences are not very pronounced when the number of testing directions is greater than or equal to 40. Therefore, we recommend using our model for NODDI parameter estimation when the total number of directions is 40 or more to ensure good results. This means that after training, our model can achieve a 6.75 $\times$ acceleration in q-space sampling while preserving good performance.

It is important to note that when the number of samples changes, all other deep learning models cannot be applied. In contrast, our method can be applied to different sampling schemes and performs significantly better than AMICO.
\begin{table*}[ht]
\centering
\caption{Quantitative comparison of SamRobNODDI under different training loss.} 
\label{tab3}
\resizebox{\textwidth}{!}{
\tiny 
\begin{tabular}{cc|cccc|cclll}
\hline                                                 
\multirow{2}{*}{Training loss}                                           &              \multirow{2}{*}{Test sampling} & \multicolumn{4}{c|}{PSNR}                  & \multicolumn{4}{c}{SSIM}               \\   \cline{3-10}  &                   & $V_{ic}$      & $V_{iso}$     & OD       & All      & $V_{ic}$      & \multicolumn{1}{c}{$V_{iso}$}    & \multicolumn{1}{c}{OD}      & \multicolumn{1}{c}{All}     \\ \hline

\multirow{2}{*}{$\mathit{L}_{r}$}                                                       
& SS                            &  32.18$\pm$ 0.65               & 36.66   $\pm$ 0.71              & 30.01    $\pm$ 0.48           & 32.17   $\pm$ 0.51              & 0.9876  $\pm$ 0.0018 & 0.9899   $\pm$ 0.0019               & 0.9816     $\pm$ 0.0024           & 0.9864  $\pm$ 0.0018 \\

& RS                             & 32.16    $\pm$ 0.64            & 36.672     $\pm$ 0.68           & 29.96     $\pm$ 0.48          & 32.13    $\pm$ 0.51      & 0.9871 $\pm$ 0.0019 & 0.9899  $\pm$ 0.0019         & 0.9813    $\pm$ 0.0024      & 0.9861  $\pm$ 0.0019 \\ \hline

\multirow{2}{*}{$\mathit{L}_{u}$}  
& SS  & 32.49  $\pm$ 0.64        & 37.23      $\pm$ 0.65           & 30.30     $\pm$ 0.49         & 32.48   $\pm$ 0.51             & 0.9884 
 $\pm$ 0.0017  & 0.9911    $\pm$ 0.0017             & 0.9828    $\pm$ 0.0023            & 0.9874    $\pm$ 0.0017                       \\

& RS    & 31.68    $\pm$ 0.61           & 36.09   $\pm$ 0.62             & 29.69   $\pm$ 0.47          & 31.76    $\pm$ 0.50            & 0.9856 $\pm$ 0.0020  & 0.9886    $\pm$ 0.0021              & 0.9800    $\pm$ 0.0025            & 0.9847  $\pm$ 0.0020   \\ \hline

\multirow{2}{*}{$\mathit{L}_{r}$+$\mathit{L}_{u}$}                                                    & SS       & 32.47      $\pm$ 0.70        & 37.34   $\pm$ 0.66              & 30.40      $\pm$ 0.50        & 32.55   $\pm$ 0.53             & 0.9885  $\pm$ 0.0016  & 0.9911   $\pm$ 0.0016              & \textbf{0.9836}    \textbf{$\pm$ 0.0022}            & 0.9877  $\pm$ 0.0017 \\

& RS        & 32.11     $\pm$ 0.69          & 36.81   $\pm$ 0.70           & 30.14   $\pm$ 0.50           & 32.23    $\pm$ 0.53            & 0.9873  $\pm$ 0.0018 & 0.9901  $\pm$ 0.0018               & 0.9820   $\pm$ 0.0024            & 0.9865  $\pm$ 0.0018  \\ \hline

\multirow{2}{*}{$\mathit{L}_{consis}$=$\mathit{L}_{r}$+$\mathit{L}_{u}$+$\mu*\mathit{L}_{r,u}$}                             
& SS   & \textbf{32.61}  \textbf{$\pm$ 0.66}              & \textbf{37.36}   \textbf{$\pm$ 0.64}               & \textbf{30.41}   \textbf{$\pm$ 0.45}             & \textbf{32.61}   \textbf{$\pm$ 0.48}               & \textbf{0.9887}  \textbf{$\pm$ 0.0016}     & \textbf{0.9913}   \textbf{$\pm$ 0.0016}                & 0.9834   $\pm$ 0.0021            & \textbf{0.9878}     \textbf{$\pm$ 0.0016}       \\
& RS                             & \textbf{32.22}     \textbf{$\pm$ 0.65}            & \textbf{36.81}    \textbf{$\pm$ 0.66}              & \textbf{30.15}      \textbf{$\pm$ 0.45}          & \textbf{32.27}    \textbf{$\pm$ 0.49}              & \textbf{0.9875}   \textbf{$\pm$ 0.0017}     & \textbf{0.9902}   \textbf{$\pm$ 0.0018}                & \textbf{0.9822}    \textbf{$\pm$ 0.0022}             & \textbf{0.9866}   \textbf{$\pm$ 0.0017}                 
  \\ \hline
\end{tabular}}
\end{table*}

\subsection{ABLATION STUDY}
To validate the effectiveness of the proposed consistency loss used in our method and further verify the model's flexibility and robustness to diffusion directions, we conducted three ablation studies: sampling consistency loss, flexible sampling scheme designs, various backbone networks.

\subsubsection{Sampling Consistency Loss}
To validate the effectiveness of the proposed consistency loss, we compared the test results of SamRobNODDI under four different training loss settings, as shown in Table \ref{tab3}. It can be observed that the results from the fourth setting, which uses sampling consistency loss, yield the best PSNR and SSIM metrics in both SS and RS testing. This indicates that the model's performance and generalization are best when utilizing the sampling consistency loss. Additionally,
the parameter $\mu$ can be adjusted for different models in the experiments. As mentioned in the Implementation Details, we set $\mu$ to 0.001. Setting $\mu$ to 0 corresponds to the third case, $\textit{L}_{r}+\textit{L}_{u}$, in Table \ref{tab3}.
\begin{figure*}[!t]
\centerline{\includegraphics[width=0.75\textwidth]{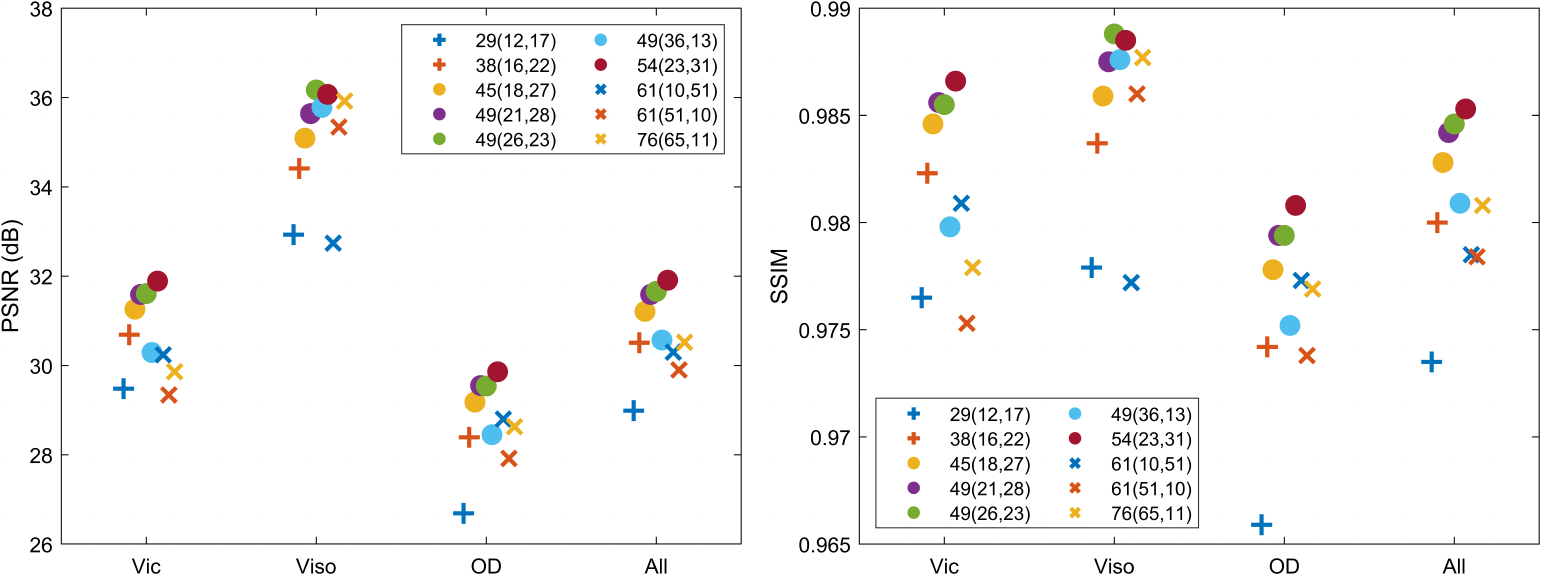}}
\caption{Quantitative comparison of SamRobNODDI under flexible sampling scheme designs in q-space. }
\label{fig6}
\end{figure*} 
\subsubsection{Flexible Sampling Scheme Designs}

We further tested the previously trained model using ten different sampling schemes: 29 (12,17), 38 (16,22), 45 (18,27), 49 (21,28), 49 (26,23), 49 (36,13), 54 (23,31), 61 (10,51), 61 (51,10), and 76 (65,11). The number before the parentheses represents the total number of directions used, while the numbers inside the parentheses represent the number of directions used for b=1000 $s/{mm}^2$ and b=2000 $s/{mm}^2$, respectively. The experimental results are shown in Fig. \ref{fig6}. To differentiate the results of different numbers of directions, we used different colors and markers (plus and cross signs) for the first two and last three samples.

From the figure, we can observe the following:

 (1) When the total number of testing directions exceeds 38, and the number of directions for a single b-value is greater than 15, the model is able to estimate parameters well. However, when the total number of directions is less than 30, or the number of directions for a single b-value is only 10, the performance is relatively poor.

 (2) Comparing the results of 29 (12,17), 61 (10,51), and 61 (51,10), we can see that the $V_{iso}$ parameter is more sensitive to the number of directions used for b=1000 $s/{mm}^2$. When the number of directions for b=1000 $s/{mm}^2$ is small, increasing the total number of directions does not significantly improve the estimation performance of $V_{iso}$. However, if the total number of directions remains the same, using more b=1000 $s/{mm}^2$ directions can effectively improve the performance of $V_{iso}$.

 (3) The comparison between 61 (10,51) and 61 (51,10) also shows that both $V_{iso}$ and OD parameters are more sensitive to b=2000 $s/{mm}^2$. With the same total number of directions, using more b=2000 $s/{mm}^2$ directions yields better results compared to using more b=1000 $s/{mm}^2$ directions.

Overall, the model can maintain optimal and stable performance when the total number of directions is greater than 40 and the number of directions for the two b-values is relatively balanced.
\begin{figure}[!t]
\centerline{\includegraphics[width=0.5\textwidth]{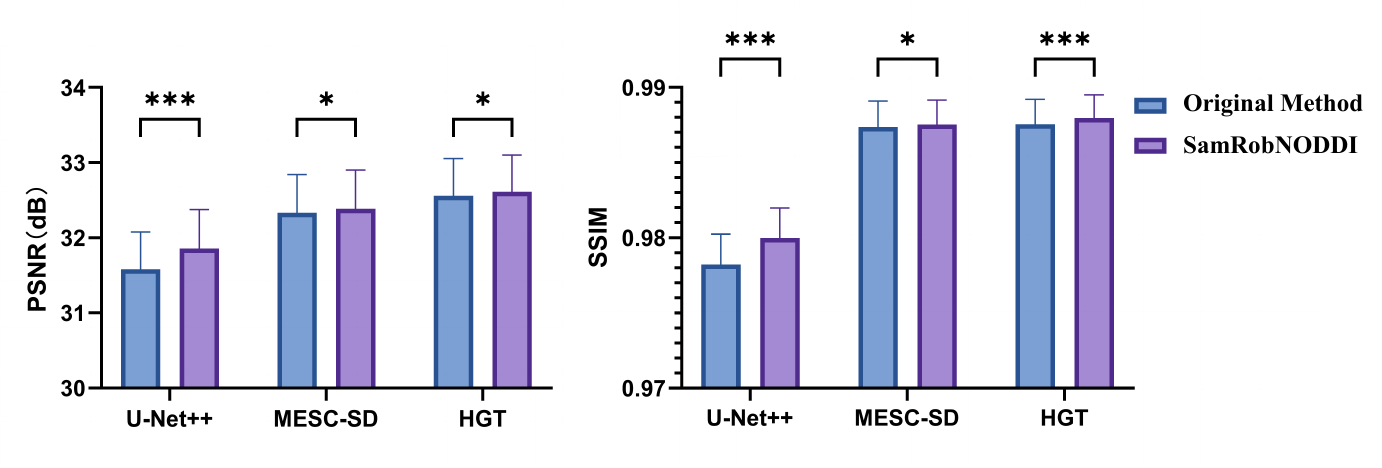}}
\caption{Quantitative comparison of SS test results for SamRobNODDI with different backbones.The p-value is described in the figure, where * represents $p<0.05$ and *** represents $p<0.001$.}
\label{fig7}
\end{figure} 
\begin{figure}[!t]
\centerline{\includegraphics[width=0.5\textwidth]{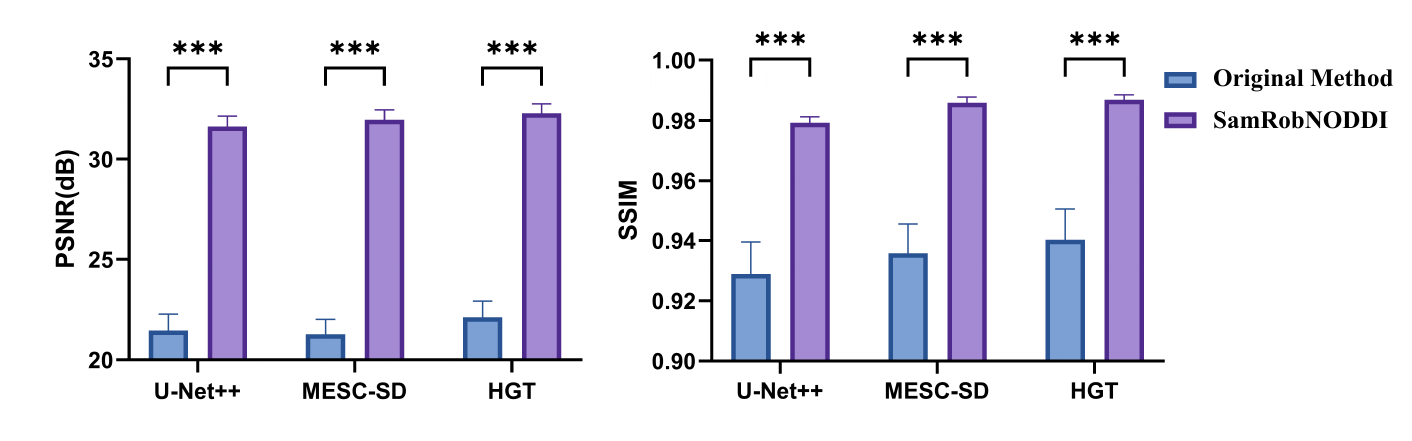}}
\caption{Quantitative comparison of RS test results for SamRobNODDI with different backbones.The p-value is described in the figure, where * represents $p<0.05$ and *** represents $p<0.001$.}
\label{fig8}
\end{figure} 
\subsubsection{Various Backbone Networks}
In the previous experiments, our method was based on an modified ResNet18 network as the backbone. However, in theory, our approach is applicable to various backbones. To verify the applicability of our method with different backbones, we selected U-Net++, MESC-SD, and HGT as backbones for training and testing. The test results are shown in Fig. \ref{fig7} and Fig. \ref{fig8}, corresponding to the quantitative metrics under SS and RS tests, respectively. As seen in Fig. \ref{fig7}, our method slightly outperforms the original methods in terms of overall performance in the SS test. In Fig. \ref{fig8}, our method significantly outperforms the original methods in the RS test. This demonstrates that our proposed approach can be used with different deep learning models as the backbone and substantially improve the robustness of the original methods to different diffusion directions.


\section{CONCLUSIONS}
\label{sec5}
In this work, we proposed SamRobNODDI to achieve robust and generalized NODDI by integrating q-space sampling augmentation, continuous representation learning, and sampling consistency loss. Experimental results indicate that the proposed method shows excellent robustness, generalization, and flexibility regarding different sampling schemes compared to existing advanced methods. Within a certain range of sampling direction numbers (e.g., when the total number of directions reaches 40 or more, and the number of directions for a single-shell is no less than 15), our trained model consistently achieves stable test results. The experiments also show that our method is not limited to specific backbones and can achieve high-quality estimation results with different backbones, significantly enhancing the robustness of the original methods. This further validates the flexibility and effectiveness of the proposed method.






\bibliography{main}   

\begin{thebibliography}{10}
\providecommand{\url}[1]{#1}
\csname url@samestyle\endcsname
\providecommand{\newblock}{\relax}
\providecommand{\bibinfo}[2]{#2}
\providecommand{\BIBentrySTDinterwordspacing}{\spaceskip=0pt\relax}
\providecommand{\BIBentryALTinterwordstretchfactor}{4}
\providecommand{\BIBentryALTinterwordspacing}{\spaceskip=\fontdimen2\font plus
\BIBentryALTinterwordstretchfactor\fontdimen3\font minus \fontdimen4\font\relax}
\providecommand{\BIBforeignlanguage}[2]{{%
\expandafter\ifx\csname l@#1\endcsname\relax
\typeout{** WARNING: IEEEtran.bst: No hyphenation pattern has been}%
\typeout{** loaded for the language `#1'. Using the pattern for}%
\typeout{** the default language instead.}%
\else
\language=\csname l@#1\endcsname
\fi
#2}}
\providecommand{\BIBdecl}{\relax}
\BIBdecl

\bibitem{le1986mr}
D.~Le~Bihan, E.~Breton, D.~Lallemand, P.~Grenier, E.~Cabanis, and M.~Laval-Jeantet, ``\textrm{MR imaging of intravoxel incoherent motions: application to diffusion and perfusion in neurologic disorders.}'' \emph{Radiology}, vol. 161, no.~2, pp. 401--407, 1986.

\bibitem{le2001diffusion}
D.~Le~Bihan, J.-F. Mangin, C.~Poupon, C.~A. Clark, S.~Pappata, N.~Molko, and H.~Chabriat, ``\textrm{Diffusion tensor imaging: concepts and applications},'' \emph{Journal of Magnetic Resonance Imaging: An Official Journal of the International Society for Magnetic Resonance in Medicine}, vol.~13, no.~4, pp. 534--546, 2001.

\bibitem{steven2014diffusion}
A.~J. Steven, J.~Zhuo, and E.~R. Melhem, ``\textrm{Diffusion kurtosis imaging: an emerging technique for evaluating the microstructural environment of the brain},'' \emph{American Journal of Roentgenology}, vol. 202, no.~1, pp. W26--W33, 2014.

\bibitem{bennett2003characterization}
K.~M. Bennett, K.~M. Schmainda, R.~Bennett, D.~B. Rowe, H.~Lu, and J.~S. Hyde, ``Characterization of continuously distributed cortical water diffusion rates with a stretched-exponential model,'' \emph{Magnetic Resonance in Medicine: An Official Journal of the International Society for Magnetic Resonance in Medicine}, vol.~50, no.~4, pp. 727--734, 2003.

\bibitem{le1988separation}
D.~Le~Bihan, E.~Breton, D.~Lallemand, M.-L. Aubin, J.~Vignaud, and M.~Laval-Jeantet, ``\textrm{Separation of diffusion and perfusion in intravoxel incoherent motion MR imaging.}'' \emph{Radiology}, vol. 168, no.~2, pp. 497--505, 1988.

\bibitem{kaden2016multi}
E.~Kaden, N.~D. Kelm, R.~P. Carson, M.~D. Does, and D.~C. Alexander, ``Multi-compartment microscopic diffusion imaging,'' \emph{NeuroImage}, vol. 139, pp. 346--359, 2016.

\bibitem{zhang2012noddi}
H.~Zhang, T.~Schneider, C.~A. Wheeler-Kingshott, and D.~C. Alexander, ``\textrm{NODDI: practical in vivo neurite orientation dispersion and density imaging of the human brain},'' \emph{Neuroimage}, vol.~61, no.~4, pp. 1000--1016, 2012.

\bibitem{palombo2020sandi}
M.~Palombo, A.~Ianus, M.~Guerreri, D.~Nunes, D.~C. Alexander, N.~Shemesh, and H.~Zhang, ``\textrm{SANDI: a compartment-based model for non-invasive apparent soma and neurite imaging by diffusion MRI},'' \emph{Neuroimage}, vol. 215, p. 116835, 2020.

\bibitem{greenspan2016guest}
H.~Greenspan, B.~Van~Ginneken, and R.~M. Summers, ``Guest editorial deep learning in medical imaging: Overview and future promise of an exciting new technique,'' \emph{IEEE Transactions on Medical Imaging}, vol.~35, no.~5, pp. 1153--1159, 2016.

\bibitem{fu2020microstructural}
X.~Fu, S.~Shrestha, M.~Sun, Q.~Wu, Y.~Luo, X.~Zhang, J.~Yin, and H.~Ni, ``\textrm{Microstructural white matter alterations in mild cognitive impairment and alzheimer’s disease: study based on Neurite Orientation Dispersion and Density Imaging (NODDI)},'' \emph{Clinical Neuroradiology}, vol.~30, pp. 569--579, 2020.

\bibitem{zheng2022adaptive}
T.~Zheng, W.~Zheng, Y.~Sun, Y.~Zhang, C.~Ye, and D.~Wu, ``\textrm{An adaptive network with extragradient for diffusion MRI-based microstructure estimation},'' in \emph{International Conference on Medical Image Computing and Computer-Assisted Intervention}.\hskip 1em plus 0.5em minus 0.4em\relax Springer, 2022, pp. 153--162.

\bibitem{park2021diffnet}
J.~Park, W.~Jung, E.-J. Choi, S.-H. Oh, J.~Jang, D.~Shin, H.~An, and J.~Lee, ``\textrm{DIFFnet: diffusion parameter mapping network generalized for input diffusion gradient schemes and b-value},'' \emph{IEEE Transactions on Medical Imaging}, vol.~41, no.~2, pp. 491--499, 2021.

\bibitem{daducci2015accelerated}
A.~Daducci, E.~J. Canales-Rodr{\'\i}guez, H.~Zhang, T.~B. Dyrby, D.~C. Alexander, and J.-P. Thiran, ``\textrm{Accelerated microstructure imaging via convex optimization (AMICO) from diffusion MRI data},'' \emph{Neuroimage}, vol. 105, pp. 32--44, 2015.

\bibitem{hernandez2019using}
M.~Hernandez-Fernandez, I.~Reguly, S.~Jbabdi, M.~Giles, S.~Smith, and S.~N. Sotiropoulos, ``Using gpus to accelerate computational diffusion mri: From microstructure estimation to tractography and connectomes,'' \emph{Neuroimage}, vol. 188, pp. 598--615, 2019.

\bibitem{golkov2016q}
V.~Golkov, A.~Dosovitskiy, J.~I. Sperl, M.~I. Menzel, M.~Czisch, P.~S{\"a}mann, T.~Brox, and D.~Cremers, ``\textrm{Q-space deep learning: twelve-fold shorter and model-free diffusion MRI scans},'' \emph{IEEE Transactions on Medical Imaging}, vol.~35, no.~5, pp. 1344--1351, 2016.

\bibitem{ye2017estimation}
C.~Ye, ``Estimation of tissue microstructure using a deep network inspired by a sparse reconstruction framework,'' in \emph{Information Processing in Medical Imaging: 25th International Conference, IPMI 2017, Boone, NC, USA, June 25-30, 2017, Proceedings 25}.\hskip 1em plus 0.5em minus 0.4em\relax Springer, 2017, pp. 466--477.

\bibitem{ye2019deep}
C.~Ye, X.~Li, and J.~Chen, ``\textrm{A deep network for tissue microstructure estimation using modified LSTM units},'' \emph{Medical image analysis}, vol.~55, pp. 49--64, 2019.

\bibitem{ye2020improved}
C.~Ye, Y.~Li, and X.~Zeng, ``An improved deep network for tissue microstructure estimation with uncertainty quantification,'' \emph{Medical Image Analysis}, vol.~61, p. 101650, 2020.

\bibitem{zheng2023microstructure}
T.~Zheng, G.~Yan, H.~Li, W.~Zheng, W.~Shi, Y.~Zhang, C.~Ye, and D.~Wu, ``\textrm{A microstructure estimation Transformer inspired by sparse representation for diffusion MRI},'' \emph{Medical Image Analysis}, vol.~86, p. 102788, 2023.

\bibitem{tian2020deepdti}
Q.~Tian, B.~Bilgic, Q.~Fan, C.~Liao, C.~Ngamsombat, Y.~Hu, T.~Witzel, K.~Setsompop, J.~R. Polimeni, and S.~Y. Huang, ``\textrm{DeepDTI: High-fidelity six-direction diffusion tensor imaging using deep learning},'' \emph{NeuroImage}, vol. 219, p. 117017, 2020.

\bibitem{gibbons2019simultaneous}
E.~K. Gibbons, K.~K. Hodgson, A.~S. Chaudhari, L.~G. Richards, J.~J. Majersik, G.~Adluru, and E.~V. DiBella, ``\textrm{Simultaneous NODDI and GFA parameter map generation from subsampled q-space imaging using deep learning},'' \emph{Magnetic Resonance in Medicine}, vol.~81, no.~4, pp. 2399--2411, 2019.

\bibitem{kebiri2024deep}
H.~Kebiri, A.~Gholipour, R.~Lin, L.~Vasung, C.~Calixto, {\v{Z}}.~Krsnik, D.~Karimi, and M.~B. Cuadra, ``\textrm{Deep learning microstructure estimation of developing brains from diffusion MRI: a newborn and fetal study},'' \emph{Medical Image Analysis}, vol.~95, p. 103186, 2024.

\bibitem{chen2023deep}
G.~Chen, Y.~Hong, K.~M. Huynh, and P.-T. Yap, ``\textrm{Deep learning prediction of diffusion MRI data with microstructure-sensitive loss functions},'' \emph{Medical Image Analysis}, vol.~85, p. 102742, 2023.

\bibitem{chen2020estimating}
G.~Chen, Y.~Hong, Y.~Zhang, J.~Kim, K.~M. Huynh, J.~Ma, W.~Lin, D.~Shen, P.-T. Yap, and U.~B. C.~P. Consortium, ``Estimating tissue microstructure with undersampled diffusion data via graph convolutional neural networks,'' in \emph{International Conference on Medical Image Computing and Computer-Assisted Intervention}.\hskip 1em plus 0.5em minus 0.4em\relax Springer, 2020, pp. 280--290.

\bibitem{yang2023towards}
J.~Yang, H.~Jiang, T.~Tassew, P.~Sun, J.~Ma, Y.~Xia, P.-T. Yap, and G.~Chen, ``\textrm{Towards Accurate Microstructure Estimation via 3D Hybrid Graph Transformer},'' in \emph{International Conference on Medical Image Computing and Computer-Assisted Intervention}.\hskip 1em plus 0.5em minus 0.4em\relax Springer, 2023, pp. 25--34.

\bibitem{chen2022hybrid}
G.~Chen, H.~Jiang, J.~Liu, J.~Ma, H.~Cui, Y.~Xia, and P.-T. Yap, ``\textrm{Hybrid graph transformer for tissue microstructure estimation with undersampled diffusion MRI data},'' in \emph{International Conference on Medical Image Computing and Computer-Assisted Intervention}.\hskip 1em plus 0.5em minus 0.4em\relax Springer, 2022, pp. 113--122.

\bibitem{koppers2019spherical}
S.~Koppers, L.~Bloy, J.~I. Berman, C.~M. Tax, J.~C. Edgar, and D.~Merhof, ``Spherical harmonic residual network for diffusion signal harmonization,'' in \emph{Computational Diffusion MRI: International MICCAI Workshop, Granada, Spain, September 2018 22}.\hskip 1em plus 0.5em minus 0.4em\relax Springer, 2019, pp. 173--182.

\bibitem{sedlar2021spherical}
S.~Sedlar, A.~Alimi, T.~Papadopoulo, R.~Deriche, and S.~Deslauriers-Gauthier, ``\textrm{A spherical convolutional neural network for white matter structure imaging via dMRI},'' in \emph{Medical Image Computing and Computer Assisted Intervention--MICCAI 2021: 24th International Conference, Strasbourg, France, September 27--October 1, 2021, Proceedings, Part III 24}.\hskip 1em plus 0.5em minus 0.4em\relax Springer, 2021, pp. 529--539.

\bibitem{nath2021dw}
V.~Nath, K.~Ramadass, K.~G. Schilling, C.~B. Hansen, R.~Fick, S.~K. Pathak, A.~W. Anderson, and B.~A. Landman, ``\textrm{DW-MRI microstructure model of models captured via single-shell bottleneck deep learning},'' in \emph{Computational Diffusion MRI: International MICCAI Workshop, Lima, Peru, October 2020}.\hskip 1em plus 0.5em minus 0.4em\relax Springer, 2021, pp. 147--157.

\bibitem{xiao2024robnoddi}
T.~Xiao, J.~Cheng, W.~Fan, J.~Yang, C.~Li, E.~Dong, and S.~Wang, ``\textrm{RobNODDI: Robust NODDI Parameter Estimation with Adaptive Sampling under Continuous Representation},'' \emph{arXiv preprint arXiv:2408.01944}, 2024.

\bibitem{caranova2023systematic}
M.~Caranova, J.~F. Soares, S.~Batista, M.~Castelo-Branco, and J.~V. Duarte, ``\textrm{A systematic review of microstructural abnormalities in multiple sclerosis detected with NODDI and DTI models of diffusion-weighted magnetic resonance imaging},'' \emph{Magnetic Resonance Imaging}, 2023.

\bibitem{he2016deep}
K.~He, X.~Zhang, S.~Ren, and J.~Sun, ``Deep residual learning for image recognition,'' in \emph{Proceedings of the IEEE Conference on Vomputer Vision and Pattern Recognition}, 2016, pp. 770--778.

\bibitem{caruyer2013design}
E.~Caruyer, C.~Lenglet, G.~Sapiro, and R.~Deriche, ``\textrm{Design of multishell sampling schemes with uniform coverage in diffusion MRI},'' \emph{Magnetic Resonance in Medicine}, vol.~69, no.~6, pp. 1534--1540, 2013.

\bibitem{cheng2014designing}
J.~Cheng, D.~Shen, and P.-T. Yap, ``\textrm{Designing single-and multiple-shell sampling schemes for diffusion MRI using spherical code},'' in \emph{Medical Image Computing and Computer-Assisted Intervention--MICCAI 2014: 17th International Conference, Boston, MA, USA, September 14-18, 2014, Proceedings, Part III 17}.\hskip 1em plus 0.5em minus 0.4em\relax Springer, 2014, pp. 281--288.

\bibitem{cheng2017single}
J.~Cheng, D.~Shen, P.-T. Yap, and P.~J. Basser, ``\textrm{Single-and multiple-shell uniform sampling schemes for diffusion MRI using spherical codes},'' \emph{IEEE Transactions on Medical Imaging}, vol.~37, no.~1, pp. 185--199, 2017.

\bibitem{descoteaux2007regularized}
M.~Descoteaux, E.~Angelino, S.~Fitzgibbons, and R.~Deriche, ``\textrm{Regularized, fast, and robust analytical Q-ball imaging},'' \emph{Magnetic Resonance in Medicine: An Official Journal of the International Society for Magnetic Resonance in Medicine}, vol.~58, no.~3, pp. 497--510, 2007.

\bibitem{van2013wu}
D.~C. Van~Essen, S.~M. Smith, D.~M. Barch, T.~E. Behrens, E.~Yacoub, K.~Ugurbil, W.-M.~H. Consortium \emph{et~al.}, ``\textrm{The WU-Minn human connectome project: an overview},'' \emph{Neuroimage}, vol.~80, pp. 62--79, 2013.

\bibitem{miller2016multimodal}
K.~L. Miller, F.~Alfaro-Almagro, N.~K. Bangerter, D.~L. Thomas, E.~Yacoub, J.~Xu, A.~J. Bartsch, S.~Jbabdi, S.~N. Sotiropoulos, J.~L. Andersson \emph{et~al.}, ``\textrm{Multimodal population brain imaging in the UK Biobank prospective epidemiological study},'' \emph{Nature Neuroscience}, vol.~19, no.~11, pp. 1523--1536, 2016.

\bibitem{zhou2018unet++}
Z.~Zhou, M.~M. Rahman~Siddiquee, N.~Tajbakhsh, and J.~Liang, ``\textrm{Unet++: A nested u-net architecture for medical image segmentation},'' in \emph{Deep Learning in Medical Image Analysis and Multimodal Learning for Clinical Decision Support: 4th International Workshop, DLMIA 2018, and 8th International Workshop, ML-CDS 2018, Held in Conjunction with MICCAI 2018, Granada, Spain, September 20, 2018, Proceedings 4}.\hskip 1em plus 0.5em minus 0.4em\relax Springer, 2018, pp. 3--11.

\end{thebibliography}
\bibliographystyle{IEEEtran}
\end{document}